\documentclass[lettersize,journal]{IEEEtran}
\usepackage{amsmath,amsfonts}
\usepackage{algorithm}
\usepackage{array}
\usepackage[caption=false,font=normalsize,labelfont=sf,textfont=sf]{subfig}
\usepackage{textcomp}
\usepackage{stfloats}
\usepackage{url}
\usepackage{bm}
\usepackage{verbatim}
\usepackage{graphicx}
\usepackage{cite}
\usepackage{todonotes}
\usepackage{hyperref}
\newcommand{\modified}[1]{\textcolor{red}{#1}}
\renewcommand{\modified}[1]{#1}

\usepackage{algpseudocode}
\usepackage{amsmath}
\usepackage{xcolor}

\begin{document}

\title{ManiSkill-ViTac 2025: Challenge on Manipulation Skill
Learning With Vision and Tactile Sensing }

\author{Chuanyu Li, Renjun Dang, Xiang Li, Zhiyuan Wu, Jing Xu, Hamidreza Kasaei, Roberto Calandra, Nathan Lepora, Shan Luo, Hao Su, Rui Chen
\thanks{Corresponding author: Rui Chen chenruithu@mail.tsinghua.edu.cn}
\thanks{
}}

\maketitle

\begin{abstract}

This article introduces the ManiSkill-ViTac Challenge 2025, which focuses on learning contact-rich manipulation skills using both tactile and visual sensing. Expanding upon the 2024 challenge, ManiSkill-ViTac 2025 includes 3 independent tracks: tactile manipulation, tactile-vision fusion manipulation, and tactile sensor structure design. The challenge aims to push the boundaries of robotic manipulation skills, emphasizing the integration of tactile and visual data to enhance performance in complex, real-world tasks. Participants will be evaluated using standardized metrics across both simulated and real-world environments, spurring innovations in sensor design and significantly advancing the field of vision-tactile fusion in robotics. 
\end{abstract}


\section{Introduction}
\IEEEPARstart{M}{anipulation} is the foundation of robotics and embodied AI \cite{doi:10.1126/science.aat8414, gu2023maniskill2, tao2024maniskill3}. 
In the last few years, manipulation has witnessed significant progress thanks to breakthroughs in reinforcement learning, computer vision, and physics simulation \cite{ramesh2023physics, zeng2020tossingbot, todorov2012mujoco}. However, performance in the real world is still far from satisfactory. Robustness and stability are still inferior, especially for contact-rich scenarios. 
One reason is that most works only use vision sensing, lacking the ability to detect contact status and use that for decision making.

The ManiSkill-ViTac Challenge aims to push the boundaries of contact-rich manipulation skill learning with tactile and vision sensing. 
Building upon last year's challenge~\footnote{\url{https://ai-workshops.github.io/maniskill-vitac-challenge-2024/}}, the challenge 2025 extends the scope and includes three independent 
 tracks, each focusing on different aspects of skill learning:
\begin{itemize}
    \item \textbf{Track 1:
    Tactile Manipulation.} This track focuses on tactile-only manipulation to develop robust policies for scenarios where visual feedback is unreliable or unavailable, such as in dark environments or occluded spaces. Understanding pure tactile manipulation provides crucial insights for developing more reliable contact-rich manipulation strategies.
    \item \textbf{Track 2: Tactile-Vision Fusion Manipulation.} Since tactile data is confined to small, localized areas, incorporating visual information enables the robot to gather distal cues. This capability allows the robot to locate targets and perform more complex tasks. We anticipate that the integration of vision will complement tactile sensing, thereby enhancing task execution.
    \item \textbf{Track 3: Sensor Structure Design.} Tactile sensing heavily relies on physical contact, and the quantity and quality of information obtained are closely linked to the shape of the sensor. We encourage participants to utilize the methods we provide to develop more effective sensor designs for the silicone components of tactile sensors.
\end{itemize}

Key features of the ManiSkill-ViTac Challenge 2025 include:
\begin{itemize}
    \item \textbf{Standardized Metrics:} We use consistent evaluation metrics across all tracks to ensure fair comparison of different algorithms. In the simulation environment, participants can freely adjust their algorithms, while the evaluation code remains the same for all. In the real-world setting, the hardware platform is identical for all participants.
    \item \textbf{Successful Sim-to-Real Transfer:} Our simulation environment closely matches real-world performance, with a small sim-to-real gap. This allows participants without access to hardware to develop and refine their algorithms in simulation, effectively contributing to the advancement of tactile manipulation.
    \item \textbf{Hardware Service Platform:} We provide a standardized hardware platform for real-world evaluation to ensure consistency in physical testing.
    \item \textbf{Multiple Manipulation Tracks:} The challenge includes tracks for tactile manipulation and tactile-vision fusion manipulation, catering to diverse research interests.
    \item \textbf{Sensor Design Track:} The challenge introduces a novel track focused on designing the silicone component of tactile sensors, promoting innovation in sensor structure. The structure of tactile sensors is crucial, with numerous recent papers focusing on structural design \cite{lepora2022digitac}. This challenge aims to further advance the field of tactile sensor structure design.
\end{itemize}

To the best of our knowledge, the ManiSkill-ViTac Challenge 2024 was the first open challenge in this field. It attracted 18 teams from around the world, and the results were presented at the 5th ViTac Workshop during ICRA 2024~\footnote{\url{https://shanluo.github.io/ViTacWorkshops/vitac2024/}}. This year, we aim to enhance the challenge format based on lessons learned and expand the discipline. We hope the challenge will foster collaboration among experts in tactile sensing, computer vision, and policy learning, driving the development of advanced manipulation skill learning and embodied AI techniques.

\section{Related work}
Robotic manipulation has traditionally relied heavily on vision-based systems, which have proven highly effective in tasks such as object detection, tracking, and grasping. Numerous vision-based benchmarks and open challenges have been established to advance this field. Challenges such as UniTeam\cite{melnik2023uniteam}, HomeRobot\cite{yenamandra2023homerobot}, ARNOLD\cite{gong2023arnold}, and ManiSkill series\cite{mu2021maniskill1, gu2023maniskill2, tao2024maniskill3} focus on manipulation tasks with visual input, such as RGB-D images and point clouds. While these vision-based approaches have shown impressive results, they often face limitations in tasks requiring fine-grained contact information, particularly when visual occlusion occurs. This has led to a growing interest in tactile sensing as a complementary modality.

Tactile sensing has emerged as a complementary approach to address these challenges. Research has shown that tactile feedback significantly enhances performance in tasks that require detailed contact information, such as detecting slippage. There are also some challenges for sensor design, \textit{e.g.} IEEE International Sensors and Measurement Student 
Contest~\footnote{\scriptsize \url{https://ieee-sensors.org/ieee-international-sensors-and-measurement-student-contest}}, which encourages participants to develop creative applications in the field of sensors and measurement systems. 

Nevertheless, manipulation tasks relying solely on tactile sensing often lack sufficient spatial awareness \cite{mandil2023tactile}, making it challenging to generalize across different objects and environments. While tactile information provides valuable insights, it typically lacks the visual context needed to handle more complex tasks. Despite the growing interest in tactile fusion, there is still no comprehensive benchmark or open challenge that fairly compares different approaches in this area. Existing benchmarks focus primarily on visual input, leaving a significant gap in evaluating the potential of multi-modal sensing.

The ManiSkill-ViTac Challenge 2025 aims to fill this gap by providing a unified platform for researchers to explore and evaluate tactile vision fusion in robotic manipulation. By integrating both vision and tactile sensing into the competition, we seek to push the boundaries of policy learning in manipulation tasks and establish a standard for assessing multi-modal approaches.

\section{Platform}
\begin{figure*}[thpb]
    \centering
    \includegraphics[width=16cm]{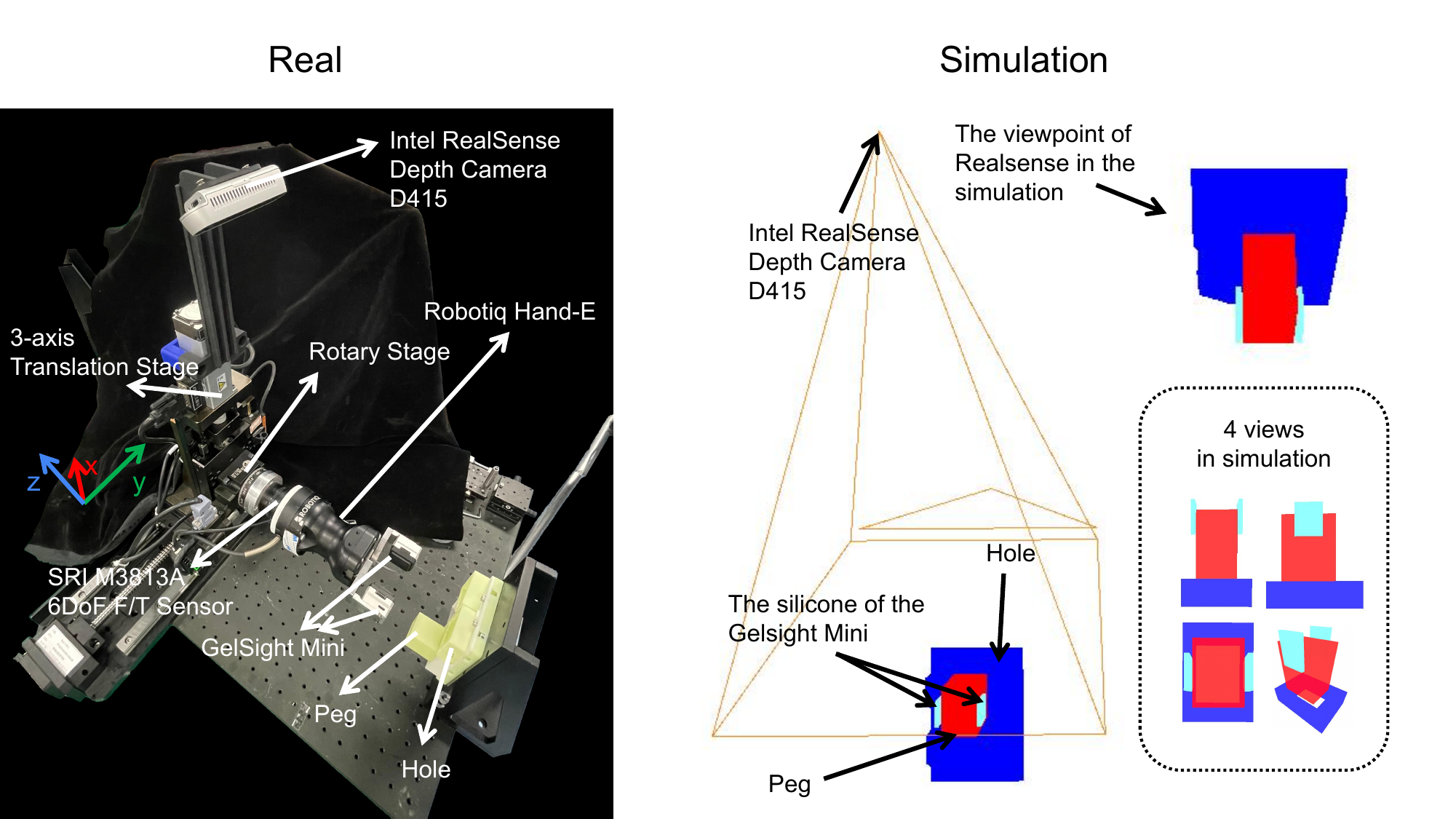}
    \caption{
    The left image shows the real-world experimental platform, consisting of a 3-axis translation stage, a rotary stage, two GelSight Mini sensors, an SRI M3813A 6DoF F/T Sensor, an Intel RealSense Depth Camera D415, and a parallel gripper (Robotiq Hand-E). The right image depicts the simulated scene, which only includes the silicone parts of the tactile sensors and the target objects for each task.
    }
    \label{fig:platform}
\end{figure*}

\subsection{Overview}
Our hardware platform consists of key components for both sensing and manipulation:
\begin{itemize}
    \item A 3-axis translation stage and rotary stage for precise positioning
    \item Dual GelSight Mini sensors for tactile feedback
    \item SRI M3813A 6DoF F/T Sensor for force/torque measurements
    \item Intel RealSense Depth Camera D415 for visual input
    \item Robotiq Hand-E parallel gripper for manipulation
\end{itemize}
The simulation platform is based on SAPIEN \cite{xiang2020sapien} and utilizes Incremental Potential Contact (IPC) \cite{li2020incremental} for simulation. Detailed information can be found in Fig. \ref{fig:platform}.

\subsection{Simulation}
\subsubsection{Tactile Sensor Simulation}

\begin{figure}[!t]
	\centering
	\includegraphics[width=8.8cm]{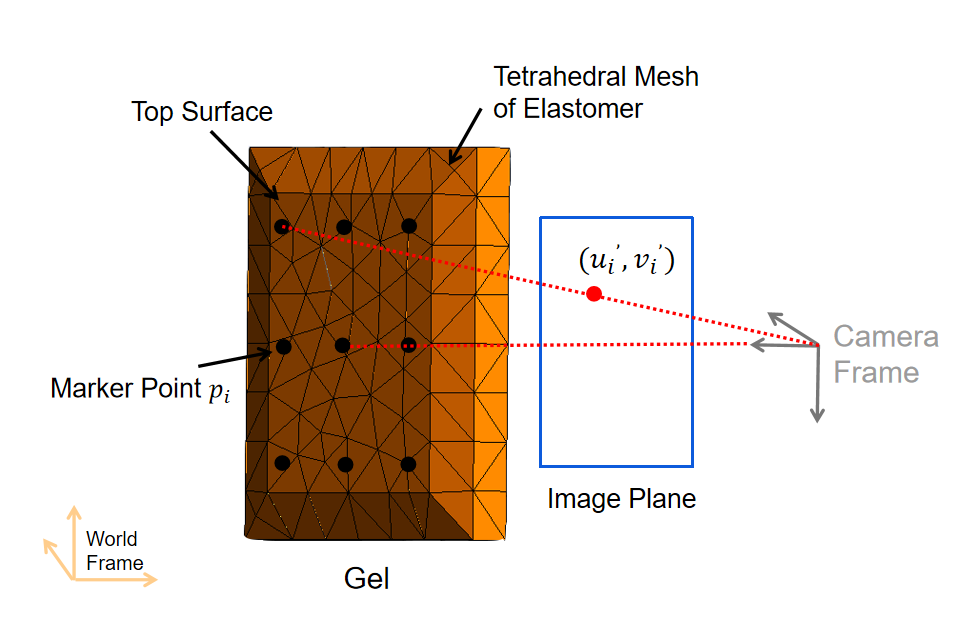}
	\caption{ 
This is a simplified GelSight Mini model, where the edge points are fixed to apply displacement. After transforming the marker point into the camera frame, its pixel coordinates can be calculated using the pinhole model. The bottom right of the figure shows an illustration of interpolation using adjacent FEM vertices. \modified{$\bm{p_i}$}  represents the $i$th marker and the facet it belongs to has 3 vertices: \modified{ $\bm{x}_{s_{i1}}$, $\bm{x}_{s_{i2}}$ and $\bm{x}_{s_{i3}}$.}}
	\label{fig:sim_gel}
\end{figure}

In the simulation, the tactile sensor’s elastomer (typically silicone rubber) is discretized using the Finite Element Method (FEM). To simulate the deformation of the elastomer, the challenge adopts the Incremental Potential Contact (IPC) method \cite{li2020incremental}, which supports hyperelastic material models such as the Neo-Hookean model. IPC is able to accurately capture the dynamic and elastic behavior of the elastomer, and more importantly, it ensures intersection-free and inversion-free simulations by employing continuous collision detection (CCD). This is crucial for stable large deformation simulations, avoiding potential errors during reinforcement learning (RL) training.

The elastomer is represented by a tetrahedral mesh, and we denote the vertices of the mesh as $\{\bm{x}_i\}$. For vertex coordinates in the world frame, we use $\{\bm{x}^W_i\}$. The markers are distributed on the top surface of the elastomer and are associated with specific facets of the mesh. Each marker $\bm{p}_i$ is located on a facet formed by three vertices $\bm{x}_{s_{i1}}, \bm{x}_{s_{i2}}, \bm{x}_{s_{i3}}$. The marker’s position can be interpolated using the barycentric weights $k_{i1}, k_{i2}, k_{i3}$:

\begin{equation}
    \bm{p}^E_i = k_{i1} \bm{x}_{s_{i1}}^E + k_{i2} \bm{x}_{s_{i2}}^E + k_{i3} \bm{x}_{s_{i3}}^E
\end{equation}

At each simulation step, IPC solves for the new positions $\{\bm{x}'^W_i\}$ of the vertices in the world frame. Subsequently, the displaced marker positions $\{\bm{p}'^W_i\}$ are calculated using the updated vertex positions and associated barycentric weights. These marker displacements are then transformed into the camera frame to generate the tactile sensor signals. Detailed information can be found in Fig. \ref{fig:sim_gel}.

\begin{figure}[!t]
	\centering
	\includegraphics[width=8.8cm]{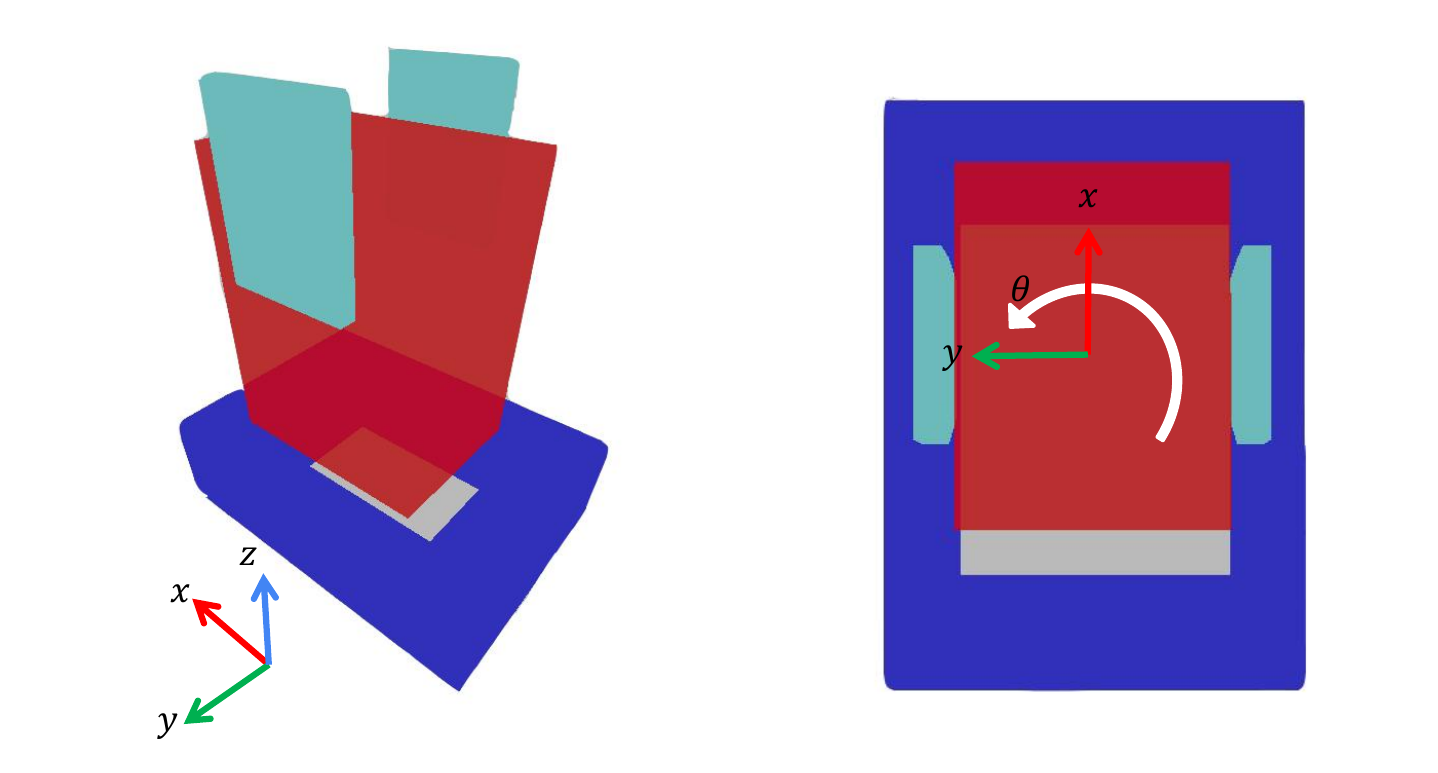}
	\caption{\modified{Frame and action direction in simulation.}}
	\label{fig:sim_action}
\end{figure}

\subsubsection{Depth Sensor Simulation}
The depth is inferred using a stereo camera in the simulation. A stereo camera system consists of one RGB camera, two IR(InfraRed) cameras and one IR projector. The position of each component is similar to the Intel RealSense Depth Camera D415. We employ Zhang's active stereo sensor simulation method\cite{10027470} to simulate the depth calculation. There are three modes to render the depth map.
\begin{enumerate}
    \item[(1)] \textbf{Ground Truth}: the depth is directly rendered as a ground truth depth map from the SAPIEN simulation environment.

    \item[(2)] \textbf{Depth Simulation in Rasterization Rendering}: enabling rasterization rendering mode in SAPIEN, the depth map is simulated using Zhang's algorithm by two IR images captured by the stereo camera.

    \item[(3)] \textbf{Depth Simulation in Ray-tracing Rendering}: enabling ray-tracing rendering mode in SAPIEN, the depth map is simulated in the same way as the second mode.
\end{enumerate}
Ray-tracing rendering increases the fidelity of IR images captured by the stereo camera, \modified{and as a result} it will affect the calculation of depth map. The simulated or ground truth depth map is aligned with the RGB frame and shares the same resolution as the RGB image.

\subsubsection{Action}
As illustrated in Fig. \ref{fig:sim_action}, the motion of the tactile sensor is induced by the action of the robot's end effector. In the simulation, these actions are applied to the elastomer as boundary conditions. The process is as follows:

\begin{enumerate}
    \item[(1)] \textbf{Setting boundary conditions}: Part of the elastomer, such as the constrained surfaces, is fixed to the sensor's shell, while the remaining surfaces, like the marker region, are free to deform. The vertices on these constrained surfaces constitute a set denoted as $\mathbb{S}$. The robot's actions dictate the motion of the elastomer, which is simulated by applying Dirichlet boundary conditions to the vertices within $\mathbb{S}$.

    \item[(2)] \textbf{Computing boundary vertex motion}: The robot’s end effector moves with a linear velocity $\bm{v}$ and an angular velocity $\omega$. Let the angular motion be defined by its direction $\bm{d}$ and a point $\bm{x}_{\text{pivot}}$. For each boundary vertex $i \in \{i | \bm{x}_i \in \mathbb{S} \}$, the position $\bm{x}'_i$ and velocity $\bm{v}_i$ at the next simulation step are computed as follows:

    \begin{equation}
    \begin{aligned}
    \bm{x}'_i &= \bm{v}\Delta t + \mathbf{R}(\bm{d}, \omega \Delta t)(\bm{x}_i - \bm{x}_{\text{pivot}}) + \bm{x}_{\text{pivot}}, \\
    \bm{v}_i &= \frac{\bm{x}'_i - \bm{x}_i }{\Delta t}
    \end{aligned}
    \end{equation}

    Here, $\Delta t$ is the timestep of the simulation and $\mathbf{R}(\bm{d}, \omega \Delta t)$ is the rotation matrix defined by axis $\bm{d}$ and rotation angle $\omega \Delta t$.
\end{enumerate}

Once the motions of the boundary vertices are determined, the FEM solver within IPC updates the entire elastomer mesh. This process enables the accurate simulation of the sensor's deformation and its interaction with objects. For detailed information, refer to Fig. \ref{fig:sim_action}.

\subsection{Real Platform}
\subsubsection{Tactile Sensor}
We have developed a real-world tactile sensing system based on the GelSight Mini sensor, utilizing the Robot Operating System (ROS) for data acquisition and communication. This system is capable of tracking the movement of surface markers in real-time and detecting contact events, thereby providing reliable sensory data for tactile interaction analysis. The system comprises the following key modules:

\paragraph{Marker Tracking}
The system receives marker tracking data from the GelSight Mini sensor via ROS topics. The sensor detects markers on its surface and provides both the initial positions and their displacements over time. By analyzing this data, the system continuously monitors the dynamic surface deformations that occur during contact, enabling effective analysis of the deformation patterns caused by interactions with objects.

\paragraph{Contact Detection}
The system determines whether the sensor is in contact with an object by monitoring the displacement of the tracked markers. When significant displacements or depth changes are detected, the system classifies it as a contact event. By comparing the movements of multiple markers, the system can also distinguish between normal contact and excessive force application. This module provides timely feedback, allowing users to accurately understand the state of the interaction.

\paragraph{System Workflow}
The system operates within the ROS framework, continuously listening to topics to receive tracking data and contact information from the GelSight Mini sensor. The core of the system is real-time data updates, ensuring that researchers can efficiently and reliably monitor the sensor's status and surface deformations. This workflow design enables the system to perform effectively in both experimental scenarios and real-world tactile sensing tasks.

\subsubsection{Depth Sensor}
We utilize the Intel RealSense Depth Camera D415 to capture aligned color and depth images, employing the Segment Anything Model~(SAM)\cite{kirillov2023segment} for image segmentation. During segmentation, masks are generated based on specific points, such as pegs and holes middle points, to extract regions of interest. These masks are then applied to the depth image, providing a clear representation of object locations. Subsequently, depth information is converted into point cloud data using the camera’s intrinsic parameters, yielding comprehensive observation data that includes both images and point clouds.

\subsubsection{Action}
The experimental platform offers four degrees of freedom in movement. The 3-axis translation stage enables translation in the x, y, and z directions, while the rotary stage provides rotation. For detailed information, refer to \modified{Fig. \ref{fig:platform}.}

\section{Format}
\begin{figure}[!t]
\centering
\includegraphics[width=8.8cm]{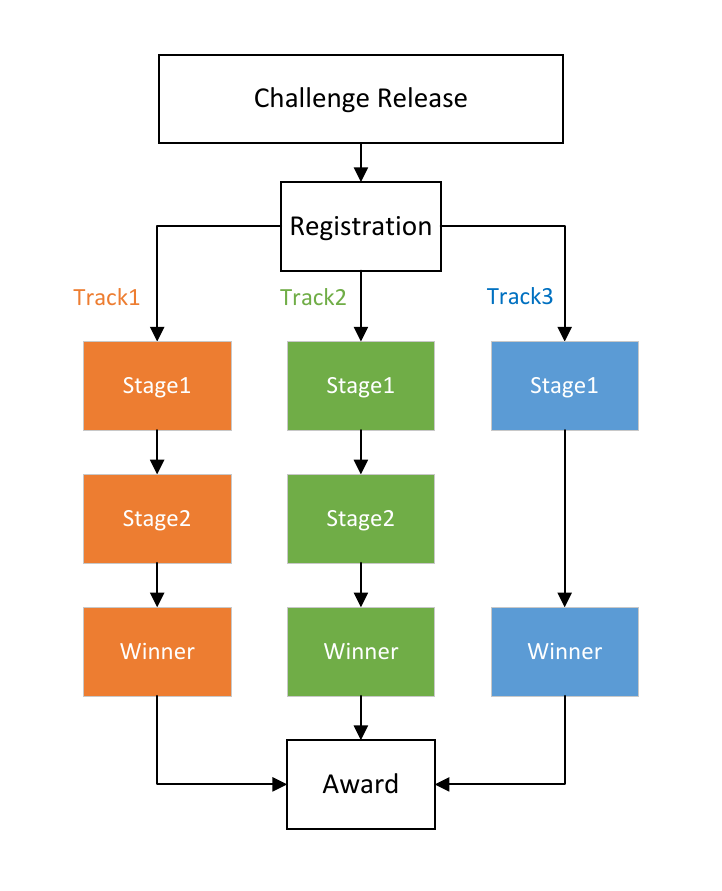}
\caption{The challenge workflow illustrates the progression through three independent tracks. Each track commences with Stage 1. Tracks 1 and 2 feature an additional Stage 2, where the number of teams advancing to this stage is determined by the total number of registrations. Track 3, on the other hand, proceeds directly from Stage 1 to the winner selection. At the conclusion of the challenge, winners from all tracks are recognized and awarded.}
\label{fig:format}
\end{figure}

As depicted in Fig. \ref{fig:format}, the challenge comprises three independent tracks, each with its own progression path. After registration, participants can compete in Tracks 1 and 2 through two stages before reaching the final winner phase. The number of teams advancing from Stage 1 to Stage 2 will be determined based on the total number of registered participants. Track 3 offers a direct progression from Stage 1 to the winner determination. Ultimately, winners from all tracks converge at the final award stage.

\subsection{Track 1} \label{subsec:Track 1}

In this track, participants are challenged with two tasks: \textbf{Peg Insertion} and \textbf{Lock Opening}. For the Peg Insertion task, there are three pairs of shapes designed to correspond to pegs and holes. The robot must accurately insert the peg into the corresponding hole, relying solely on tactile information. In the Lock Opening task, there are four pairs of shapes that correspond to keys and locks. The robot is required to accurately insert the key into the lock hole, utilizing tactile information for this precision task.
\subsubsection{Observation}
The observation design consists of two key components:

\begin{itemize}
    \item \textbf{Relative Motion (relative\_motion)}: This vector represents the sensor's current offset in terms of x, y, and z coordinates, and the rotation angle theta relative to its initial position and orientation. It provides crucial information about the sensor's movement and serves as input into the actor network to guide decisions throughout task execution. This vector tracks the sensor's movement and rotation since the task began.

\item \textbf{Marker Flow (marker\_flow)}: The marker flow captures the motion of specific marker points on the surface, which are tracked by tactile sensors. The flow is represented by two arrays: the initial marker points and the current marker points. These arrays are processed to account for noise and tracking losses, resulting in a flow of marker points that represent changes in the surface's perception by the sensors. The marker flow provides real-time feedback on surface movement and deformation, enhancing the system's understanding of sensor-surface interactions.
\end{itemize}

\subsubsection{Action}
\paragraph{Peg Insertion Task}
In this task, the precise application of actions is crucial for controlling object movements. Each action is composed of three components: $\delta_x$, $\delta_y$, and $\delta_\theta$, representing displacement increments in the $x$ and $y$ directions (in millimeters) and an angular increment around the $z$-axis (in degrees), respectively. 

To correctly apply the action in the simulation, the following steps are implemented:

First, action clipping is performed to restrict the input action within a predefined maximum allowable range, preventing excessively large actions from causing instability or non-physical behavior.

Next, since the object may have rotated, the displacement increments are converted from the object's local coordinate system to the global coordinate system. Using the current rotation angle of the object, $\theta_{\text{current}}$, the global displacement increments are computed through a two-dimensional rotation transformation:

\begin{equation}
\begin{pmatrix}
\delta_x' \\
\delta_y'
\end{pmatrix}
=
\begin{pmatrix}
\cos\theta_{\text{current}} & -\sin\theta_{\text{current}} \\
\sin\theta_{\text{current}} & \cos\theta_{\text{current}}
\end{pmatrix}
\begin{pmatrix}
\delta_x \\
\delta_y
\end{pmatrix}
\end{equation}

Then, the transformed increments are added to the current displacement and angle offsets to update the target position and orientation:

\begin{equation}
\begin{aligned}
x_{\text{offset}} & \leftarrow x_{\text{offset}} + \delta_x' \\
y_{\text{offset}} & \leftarrow y_{\text{offset}} + \delta_y' \\
\theta_{\text{offset}} & \leftarrow \theta_{\text{offset}} + \delta_\theta
\end{aligned}
\end{equation}

To ensure simulation stability, the total action is divided into multiple substeps. Based on the simulation time step $\Delta t$, the required number of substeps $N_{\text{sub}}$ is calculated to keep the displacement and rotation in each substep within allowable limits:

\begin{equation}
N_{\text{sub}} = \max\left(1,\ \left\lceil \dfrac{\max\left(|\delta_x'|,\ |\delta_y'|\right)}{v_{\text{max}} \Delta t} \right\rceil,\ \left\lceil \dfrac{|\delta_\theta|}{\omega_{\text{max}} \Delta t} \right\rceil \right)
\end{equation}

where $v_{\text{max}}$ and $\omega_{\text{max}}$ are the predefined maximum linear and angular velocities.

The linear and angular velocities applied in each substep are then calculated:

\begin{equation}
\begin{aligned}
v_x &= \dfrac{\delta_x'}{N_{\text{sub}} \Delta t} \\
v_y &= \dfrac{\delta_y'}{N_{\text{sub}} \Delta t} \\
\omega &= \dfrac{\delta_\theta}{N_{\text{sub}} \Delta t}
\end{aligned}
\end{equation}

In the simulation loop, these velocities are applied in each substep. The tactile sensors are assigned the linear velocities $(v_x, v_y)$ and angular velocity $\omega$, and the physics engine advances the simulation by one time step $\Delta t$.

After each substep, the object's displacement and angle are checked to ensure they do not exceed the allowable maximum offsets:

\begin{equation}
\begin{aligned}
|x_{\text{offset}}| & \leq x_{\text{max}} \\
|y_{\text{offset}}| & \leq y_{\text{max}} \\
|\theta_{\text{offset}}| & \leq \theta_{\text{max}}
\end{aligned}
\end{equation}

If any exceedance is detected, an error flag is set, and appropriate handling is performed.

By following these steps, actions are precisely applied in the simulation environment, ensuring that the object moves along the expected trajectory while maintaining simulation stability and accuracy. The details are shown at Algorithm \ref{alg:simstep_peg_track1}.

\begin{algorithm}
\caption{Simulation Step For Peg Insertion} \label{alg:simstep_peg_track1}
\begin{algorithmic}[1]
\Function{SimStep}{action}
    \State action $\gets$ clip(action, -max\_action, max\_action)
    \State Calculate action\_x, action\_y, action\_theta
    \State action\_sim $\gets$ [action\_x, action\_y, action\_theta]
    \State num\_substeps $\gets$ CalculateSubsteps(action\_sim)
    \For{$i = 1$ to num\_substeps}
        \State ApplyVelocityToSensors(action\_sim, num\_substeps)
        \State StepIPCSystem()
        \State StepTactileSensors()
    \EndFor
    \State z\_substeps $\gets$ CalculateZSubsteps(action\_z)
    \For{$i = 1$ to z\_substeps}
        \State ApplyZVelocityToSensors(action\_z, z\_substeps)
        \State StepIPCSystem()
        \State StepTactileSensors()
    \EndFor
\EndFunction
\end{algorithmic}
\end{algorithm}

\paragraph{Lock Opening Task}
In this task, actions are applied to control object movements in the $x$, $y$, and $z$ directions. Each action consists of displacement increments in these three directions. The following steps describe the process:

First, the number of substeps is calculated to divide the total action into smaller, manageable parts. This is based on the maximum absolute value of the action and the simulation time step $\Delta t$:

\begin{equation}
N_{\text{sub}} = \max\left(1, \left\lceil \dfrac{\max\left(\left|\delta_x\right|, \left|\delta_y\right|, \left|\delta_z\right|\right)}{2 \times 10^{-3} \times \Delta t} \right\rceil \right)
\end{equation}

Next, the velocities in each direction ($x$, $y$, and $z$) are computed by dividing the displacement by the total number of substeps and the simulation time step:

\begin{equation}
\begin{aligned}
v_x &= \dfrac{\delta_x}{N_{\text{sub}} \times \Delta t} \\
v_y &= \dfrac{\delta_y}{N_{\text{sub}} \times \Delta t} \\
v_z &= \dfrac{\delta_z}{N_{\text{sub}} \times \Delta t}
\end{aligned}
\end{equation}

During each substep, these velocities are applied to the tactile sensors to ensure smooth and controlled movement.
The details are shown at Algorithm \ref{alg:simstep_key_track1}.

\begin{algorithm}
\caption{Simulation Step for Lock Opening} \label{alg:simstep_key_track1}
\begin{algorithmic}[1]
\Function{SimStep}{action}
    \State Calculate substeps and velocity
    \For{$i = 1$ to substeps}
        \State ApplyVelocityToSensors(velocity)
        \State StepIPCSystem()
        \State StepTactileSensors()
    \EndFor
\EndFunction

\end{algorithmic}
\end{algorithm}

\subsubsection{Evaluation Metric}
\paragraph{Peg Insertion Task}  
The evaluation of task success is based on multiple criteria to ensure both accurate tracking and robust error handling. 
\begin{itemize}
    \item \textbf{Excessive Error Check:} The task fails if the peg's position or orientation exceeds defined spatial boundaries. This ensures the peg remains within the intended trajectory throughout the task.

    \item \textbf{Step Limit:} To promote efficiency, a maximum step limit is set per episode. If this limit is reached before task completion, the episode fails, as prolonged or stalled task progress is discouraged. This constraint helps ensure timely task execution within a reasonable step count.

    \item \textbf{Depth and z-Axis Translation:} Successful insertion requires the peg to descend to a minimum depth along the z-axis. This depth is calculated as the product of the number of steps taken and the z-axis step size, confirming the peg’s progression toward the target depth. Meeting this criterion is essential for successful insertion.

    \item \textbf{Surface Point Consistency:} The final success criteria involve comparing the observed surface points from two tactile sensors with their initial reference positions. The Euclidean distances between these current and initial points are calculated for both left and right surfaces. If the mean distances (\texttt{l\_diff} and \texttt{r\_diff}) are below a predefined threshold, this indicates accurate alignment and stability, confirming the peg is well-positioned.
\end{itemize}

This evaluation metric provides a comprehensive assessment by integrating movement constraints with precise surface tracking, ensuring both accuracy and reliability in task execution.

\paragraph{Lock Opening Task}
The evaluation of task success is based on multiple criteria to ensure both accurate tracking and robust error handling. 

\begin{itemize}
    \item \textbf{Excessive Error Check:} The task fails if the peg's position or orientation exceeds defined spatial boundaries. This ensures the peg remains within the intended trajectory throughout the task.

    \item \textbf{Step Limit:} To promote efficiency, a maximum step limit is set per episode. If this limit is reached before task completion, the episode fails, as prolonged or stalled task progress is discouraged. This constraint helps ensure timely task execution within a reasonable step count.

    \item \textbf{Key and Lock Alignment:} Successful insertion requires precise alignment between each pair of key and lock teeth and grooves. This alignment is confirmed by ensuring that the xyz coordinate error for each pair is below a specified threshold, indicating accurate positioning of the key within the lock. Meeting this criterion is essential for successful insertion.

    \item \textbf{Surface Point Consistency:} The final success criteria involve comparing the observed surface points from two tactile sensors with their initial reference positions. The Euclidean distances between these current and initial points are calculated for both left and right surfaces. If the mean distances (\texttt{l\_diff} and \texttt{r\_diff}) are below a predefined threshold, this indicates accurate alignment and stability, confirming the peg is well-positioned.
\end{itemize}

This evaluation metric provides a comprehensive assessment by integrating movement constraints with precise surface tracking, ensuring both accuracy and reliability in task execution.

\subsubsection{Submission}
\begin{itemize}
   \item \textbf{Stage 1 Submission:}
   \begin{itemize}
       \item Submit evaluation logs for both tasks to our email
       \item Subject line format: teamname-Track\_1-stage1-result
       \item Multiple submissions are allowed
   \end{itemize}
   
   \item \textbf{Stage 2 Submission:} (For qualified teams only)
   \begin{itemize}
       \item Submit a compressed folder named ``Track\_1" containing:
       \begin{itemize}
           \item Modified network architecture
           \item Your best performing model
       \end{itemize}
       \item Subject line format: teamname-Track\_1-stage2-result
       \item Maximum of three submissions allowed
       \item Submission deadline: before Stage 2 ends
   \end{itemize}
\end{itemize}

\subsection{Track 2}
In this track, the environment is designed with two distinct types of holes, and the robot must select the correct hole to insert the peg, which is grasped in the gripper, based on visual information provided by the system.
\subsubsection{Observation}
The observation design consists of five key components:
\begin{itemize}

    \item \textbf{Relative Motion (relative\_motion)}: This vector represents the sensor’s current offset in terms of x, y, z coordinates, and the rotation angle theta relative to its initial position and orientation. It provides crucial information about the sensor’s movement. This vector tracks how far the sensor has moved and how much it has rotated since the task began.
    
    \item \textbf{Marker Flow (marker\_flow)}: Same to Track 1.

    \item \textbf{RGB Picture (rgb\_picture)}: This component records the RGB image data captured by the depth sensor at the current observation. 

    \item \textbf{Depth Picture (depth\_picture)}: This component records the depth image data captured by the depth sensor at the current observation. 

    \item \textbf{Point Cloud (point\_cloud)}: This component records the point cloud data captured by the depth sensor at the current observation. 

\end{itemize}
For the three types of vision information, participants can choose one or more of them depending on the network they use

\subsubsection{Action}
In this task, precise control over object movements is implemented through action components \(\delta_x\), \(\delta_y\), \(\delta_\theta\), and \(\delta_z\), which correspond to displacements in the \(x\), \(y\), and \(z\) directions (in millimeters) and an angular increment around the \(z\)-axis (in degrees), respectively. The following steps describe the logic implemented to ensure stable simulation behavior and accurate action application.

First, action clipping restricts the input action within a predefined maximum range, preventing excessively large displacements or rotations that could lead to instability.

The displacement increments in \(x\) and \(y\) directions are then adjusted to account for the current orientation of the object. Using the angle \(\theta_{\text{current}}\) (converted to radians), the increments are rotated from the local to the global coordinate system:

\begin{equation}
\begin{pmatrix}
\delta_x' \\
\delta_y'
\end{pmatrix}
=
\begin{pmatrix}
\cos\theta_{\text{current}} & -\sin\theta_{\text{current}} \\
\sin\theta_{\text{current}} & \cos\theta_{\text{current}}
\end{pmatrix}
\begin{pmatrix}
\delta_x \\
\delta_y
\end{pmatrix}
\end{equation}

These transformed values, along with \(\delta_\theta\) and \(\delta_z\), update the current offsets for \(x\), \(y\), \(\theta\), and \(z\):

\begin{equation}
\begin{aligned}
x_{\text{offset}} & \leftarrow x_{\text{offset}} + \delta_x' \\
y_{\text{offset}} & \leftarrow y_{\text{offset}} + \delta_y' \\
\theta_{\text{offset}} & \leftarrow \theta_{\text{offset}} + \delta_\theta \\
z_{\text{offset}} & \leftarrow z_{\text{offset}} + \delta_z
\end{aligned}
\end{equation}

For stability, the action is divided into substeps. The required substeps \(N_{\text{sub}}\) are calculated to ensure the displacement and rotation increments per substep stay within allowed limits, given the simulation time step \(\Delta t\):

\begin{equation}
N_{\text{sub}} = \max\left(1, \ 
\begin{aligned}
&\left\lceil \frac{\max(|\delta_x'|, |\delta_y'|, |\delta_z|)}{5 \times 10^{-3} \cdot \Delta t} \right\rceil, 
&\left\lceil \frac{|\delta_\theta|}{0.2 \cdot \Delta t} \right\rceil
\end{aligned}
\right)
\end{equation}

The velocities for each substep in the \(x\), \(y\), \(z\), and \(\theta\) directions are then determined:

\begin{equation}
\begin{aligned}
v_x &= \frac{\delta_x'}{N_{\text{sub}} \cdot \Delta t} \\
v_y &= \frac{\delta_y'}{N_{\text{sub}} \cdot \Delta t} \\
v_z &= \frac{\delta_z}{N_{\text{sub}} \cdot \Delta t} \\
\omega &= \frac{\delta_\theta}{N_{\text{sub}} \cdot \Delta t}
\end{aligned}
\end{equation}

\begin{algorithm}
\caption{Simulation Step For Peg Insertion} \label{alg:simstep_peg_track2}
\begin{algorithmic}[1]
\Function{SimStep}{action}
    \State action $\gets$ clip(action, -max\_action, max\_action)
    \State Convert action to global coordinates: calculate $action\_x$, $action\_y$, $action\_theta$, and $action\_z$
    \State action\_sim $\gets$ [$action\_x$, $action\_y$, $action\_theta$, $action\_z$]
    
    \State num\_substeps $\gets$ CalculateSubsteps(action\_sim)
    \For{$i = 1$ to num\_substeps}
        \State Apply velocities ($v_x$, $v_y$, $v_z$, $v_\theta$) to tactile sensors
        \State Step IPC System
        \State Step Tactile Sensors
    \EndFor
\EndFunction
\end{algorithmic}
\end{algorithm}

This structured approach allows for controlled, realistic actions within the simulation, maintaining accuracy and stability.

\subsubsection{Evaluation Metric}
The evaluation of task success is based on several criteria, ensuring robust error handling and precise observations. Initially, the system checks for two failure conditions: excessive error (\texttt{error\_too\_large}) and an excessive number of steps (\texttt{too\_many\_steps}). If neither condition is met, the task's success is evaluated using the following metrics:

\begin{itemize}
    \item \textbf{Excessive Error Check:} The task fails if the peg's position or orientation exceeds defined spatial boundaries. This ensures the peg remains within the intended trajectory throughout the task.

    \item \textbf{Step Limit:} To promote efficiency, a maximum step limit is set per episode. If this limit is reached before task completion, the episode fails, as prolonged or stalled task progress is discouraged. This constraint helps ensure timely task execution within a reasonable step count.

    \item \textbf{Depth and z-Axis Translation:} Successful insertion requires the peg to descend to a minimum depth along the z-axis. This depth is calculated based on the relative position of the bottom of the peg to the upper surface of the hole, confirming the peg’s progression toward the target depth. Meeting this criterion is essential for successful insertion.
    
    \item \textbf{Surface Point Consistency:} The success criteria then involve comparing the observed surface points of two sensors to their initial positions. The Euclidean distances between the current and initial points are calculated for both the left and right surfaces. If the mean distances (\texttt{l\_diff} and \texttt{r\_diff}) for both surfaces are below a predefined threshold, the task is marked as successful.

\end{itemize}

\subsubsection{Submission}

\begin{itemize}
   \item \textbf{Stage 1 Submission:}
   \begin{itemize}
       \item Submit evaluation log for the task to our email
       \item Subject line format: teamname-Track\_2-stage1-result
       \item Multiple submissions are allowed
   \end{itemize}
   
   \item \textbf{Stage 2 Submission:} (For qualified teams only)
   \begin{itemize}
       \item Submit a compressed folder named ``Track\_2" containing:
       \begin{itemize}
           \item Modified network architecture
           \item Your best performing model
       \end{itemize}
       \item Subject line format: teamname-Track\_2-stage2-result
       \item Maximum of three submissions allowed
       \item Submission deadline: before Stage 2 ends
   \end{itemize}
\end{itemize}
\subsection{Track 3}

In this track, it is necessary to design the structure of the tactile sensor and complete the same task as the peg insertion in Track 1.

\subsubsection{Observation and Action}
The observation and action processes are the same as those outlined in Track 1. For more details, refer to Section \ref{subsec:Track 1}.

\subsubsection{Structure Design}
This section concentrates on optimizing the sensor's structure based on the GelSight Mini design. Participants can utilize modeling software to craft a model of the silicone component, which can then be imported into our simulation environment for verification and testing.

\subsubsection{Evaluation Metric}
The evaluation metric consists of two parts. The first part is identical to that of Track 1. The second part evaluates the innovativeness of the sensor design, focusing on both the structural design and the marker distribution.

\subsubsection{Submit}
\begin{itemize}
   \item \textbf{Stage 1 Submission:}
   \begin{itemize}
       \item Submit evaluation logs for peg insertion task to our email
       \item A documentation detailing their design methodology and implementation process
       \item Subject line format: teamname-Track\_3-stage1-result
       \item Multiple submissions are allowed
   \end{itemize}
\end{itemize}

\section{Reference Method} 

\begin{figure*}[thpb]
    \centering
    \includegraphics[width=17cm]{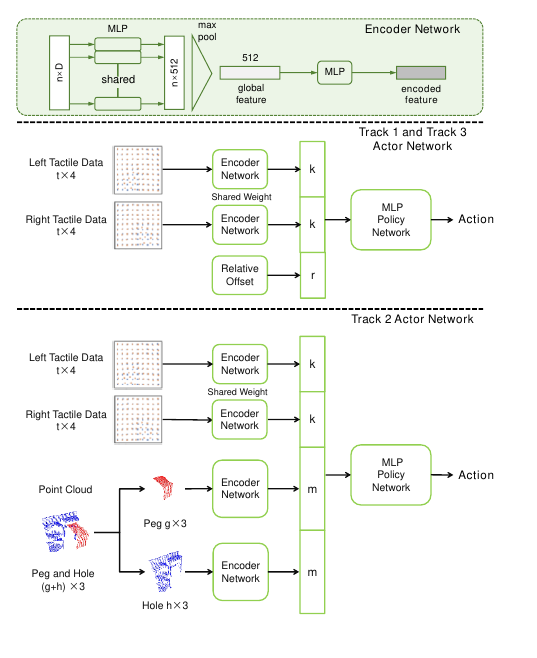}
    \caption{The diagram illustrates the actor networks for three tracks. In Tracks 1 and 3, marker flows obtained from two tactile sensors, along with the relative motion, are fed into the Encoder Network. The resulting features are concatenated and subsequently input into the MLP Policy Network to produce action outputs. In Track 2, inputs include the peg point cloud, hole point cloud, and marker flows from both left and right tactile sensors; these inputs are processed through the Encoder Network. The outputs from this network are concatenated and then input into the MLP Policy Network to generate action outputs.}
    \label{fig:rl_network}
\end{figure*}

\subsection{Track 1}
\subsubsection{Policy Network}
Referring to \cite{chen2024general}, we utilize the Twin-Delayed Deep Deterministic Policy Gradient (TD3) algorithm \cite{dankwa2019twin}. Our policy network processes input from two tactile sensors through a Shared Weight Encoder Network, which extracts $k$-dimensional feature vectors. These features, combined with the relative motion, form the input to the actor network.

At each step, the network processes:
\begin{itemize}
    \item Marker flow
    \item Relative motion
\end{itemize}

The actor network outputs a 3-dimensional action vector $[a_x, a_y, a_{\theta}]$ relative to the peg (See Fig.\ref{fig:sim_action}).

The critic network evaluates action Q-values using task-specific ground truth offsets:
\begin{itemize}
    \item For peg insertion: the offset between the peg and hole
    \item For lock opening: the offset between key teeth and pin positions
\end{itemize}

\subsubsection{Reward}

\begin{enumerate}
    \item \textbf{Peg insertion:} The reward function consists of four parts: the decrease in error $e_{t-1} - e_t$, a constant penalty $P$ for each step, the final success reward $R_{final}$, and the penalty for exceeding the allowable error $R_{fail}$. The error $e_t$ is calculated from the errors in each axis, where $e_x$ and $e_y$ are in millimeters, and $e_{\theta}$ is in degrees. To prevent the policy from failing too quickly, a penalty for large errors is applied, where $t_{max}$ is the maximum number of allowed attempts.This penalty is triggered when any positional error \( |e_{x,y}| \) exceeds a predefined threshold \(\tau_{xy}\), or when the angular error \( |e_{\theta}| \) exceeds \(\tau_{\theta}\). 

    \begin{equation}
    R_t = e_{t-1} - e_t - P + R_{\text{final}} + R_{\text{fail}}
    \end{equation}
    \begin{equation}
    e_t = \sqrt{e_x^2 + e_y^2 + e_{\theta}^2}
    \end{equation}
    \begin{equation}
    R_{\text{final}} =
    \begin{cases}
    10, & \text{if success} \\
    0, & \text{otherwise}
    \end{cases}
    \end{equation}
    \begin{equation}
    R_{\text{fail}} =
    \begin{cases}
    -2(t_{\text{max}} - t)P, & \text{if } |e_{x,y}| \geq \tau_{xy} \text{ or } |e_{\theta}| \geq \tau_{\theta} \\
    0, & \text{otherwise}
    \end{cases}
    \end{equation}
    \item \textbf{Lock Opening:} The reward function consists of four parts: the decrease in error \( e_{t-1} - e_t \), a constant penalty \( P \) for each step, the final success reward \( R_{\text{final}} \), and the penalty for exceeding the allowable error \( R_{\text{fail}} \). The error \( e_t \) is calculated from the errors in each axis, where \( e_x \), \( e_y \), and \( e_z \) are in millimeters. To prevent the policy from failing too quickly, a penalty for large errors is applied, where \( t_{\text{max}} \) is the maximum number of allowed attempts. This penalty is triggered when any positional error \( |e_{x,y}| \) exceeds a predefined threshold \( \tau_{xyz} \), or when the angular error \( |e_{\theta}| \) exceeds \( \tau_{\theta} \). Additionally, a penalty based on \( \text{surface\_diff} \) is included, which reflects the clipped deviation between observed and initial surface points, further encouraging precise alignment.

    \begin{equation}
    R_t = e_{t-1} - e_t - P + R_{\text{final}} + R_{\text{fail}}
    \end{equation}
    \begin{equation}
    e_t = 500(|e_x| + |e_y| + |e_z|)
    \end{equation}
    \begin{equation}
    R_{\text{final}} =
    \begin{cases}
    10, & \text{if success} \\
    0, & \text{otherwise}
    \end{cases}
    \end{equation}
    \begin{equation}
    R_{\text{fail}} =
    \begin{cases}
    -10(t_{\text{max}} - t)P - \sum \text{surface\_diff}, & \text{if failure} \\
    0, & \text{otherwise}
    \end{cases}
    \end{equation}
    
\end{enumerate}

\subsection{Track 2}
\subsubsection{Policy Network}
Similar to Track 1, we use the TD3 algorithm and Shared Tactile Encoder to process tactile data. For the newly added visual information, we use the SAM to segment the point clouds of peg and hole and use the Encoder Network to extract features separately. These features, combined with the features extracted from tactile data, form the input of actor network. The actor network is an MLP Network and outputs a 4-dimensional action vector $[a_x, a_y, a_z, a_{\theta}]$ relative to the peg (See Fig.\ref{fig:sim_action}).

The critic network evaluates action Q-values using task-specific ground truth offsets:
\begin{itemize}
    \item  Offset between the peg and appropriate hole
\end{itemize}

\subsubsection{Reward}
The reward function consists of four parts: the decrease in error $e_{t-1} - e_t$, a constant penalty $P$ for each step, the final success reward $R_{final}$, and the penalty for exceeding the allowable error $R_{fail}$. The error $e_t$ is calculated from the errors in each axis, where $e_x$, $e_y$ and $e_z$ are in millimeters, and $e_{\theta}$ is in degrees. To prevent the policy from failing too quickly, a penalty for large errors is used, where $t_{max}$ is the maximum number of allowed attempts. This penalty is triggered when any positional error \( |e_{x,y}| \) exceeds a predefined threshold \(\tau_{xy}\), or when the angular error \( |e_{\theta}| \) exceeds \(\tau_{\theta}\). 

\begin{equation}
R_t = e_{t-1} - e_t - P + R_{\text{final}} + R_{\text{fail}}
\end{equation}
\begin{equation}
e_t = \sqrt{e_x^2 + e_y^2 + e_z^2 +e_{\theta}^2}
\end{equation}
\begin{equation}
R_{\text{final}} =
\begin{cases}
10, & \text{if success} \\
0, & \text{otherwise}
\end{cases}
\end{equation}
\begin{equation}
R_{\text{fail}} =
\begin{cases}
-2(t_{\text{max}} - t)P, & \text{if } |e_{x,y,z}| \geq \tau_{xy} \text{ or } |e_{\theta}| \geq \tau_{\theta} \\
0, & \text{otherwise}
\end{cases}
\end{equation}

\subsection{Track 3}
\subsubsection{Policy Network and Reward}
The policy network and reward setup is the same as in Track 1.

\subsubsection{Gel Model Generation}

When designing the gel model, ensure that the base size is 20.75 mm by 25.25 mm, as this is the default size of the Gelsight Mini used in the challenge. You can create your model manually or use parametric modeling software, such as SolidWorks.

\begin{figure}[!t]
	\centering
	\includegraphics[width=6.8cm]{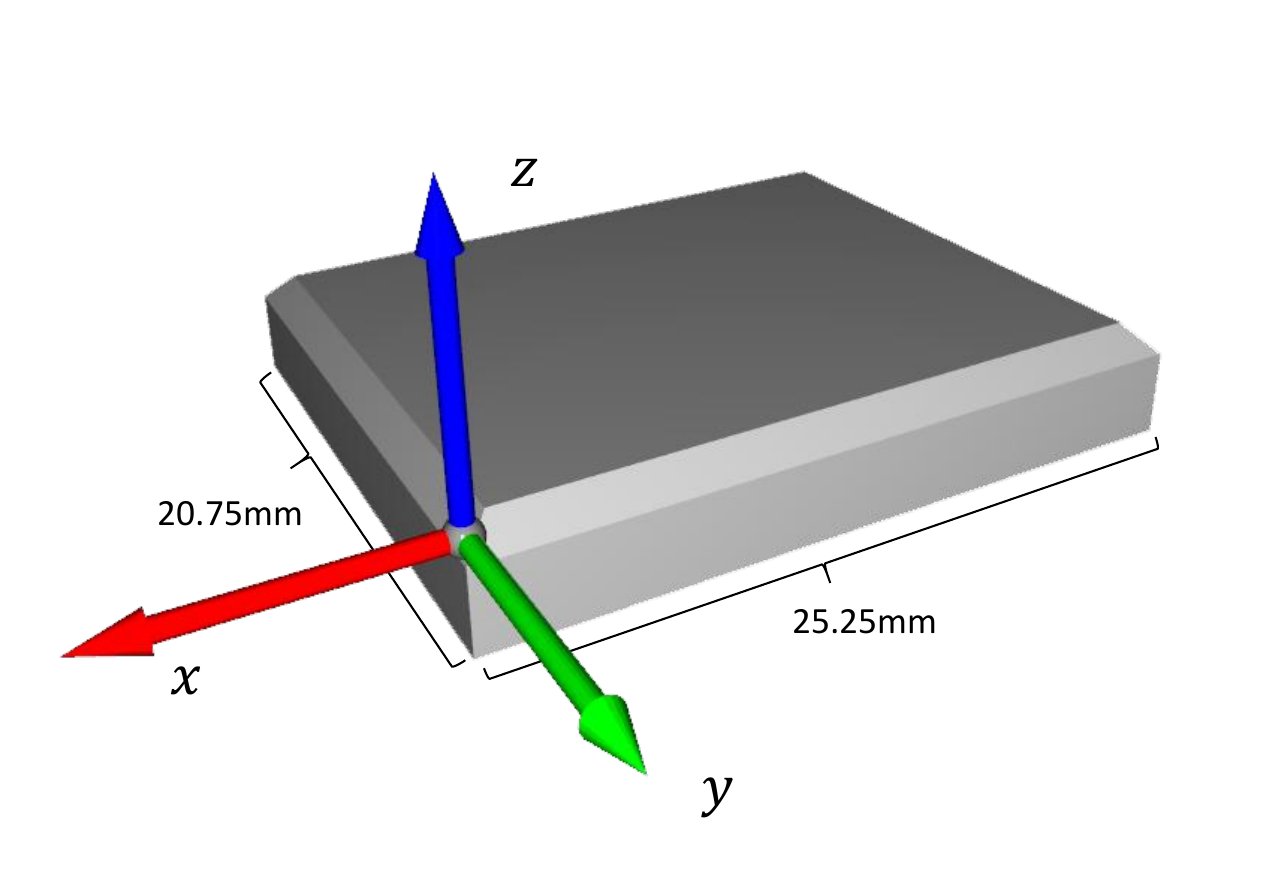}
	\caption{STL file orientation requirements for the gel model: The 25.25 mm edge must align with the x-axis, while the sensing surface orients along the positive z-axis. This alignment ensures consistent marker tracking and proper force distribution during deformation simulations.}
	\label{fig:STL_pos}
\end{figure}

\begin{enumerate}
    \item \textbf{Generate STL File:} Use modeling software to generate the STL file of the gel part you designed. Note: since we are using Gelsight Mini as the base, the part where the gel and the casing are attached should be sized at 20.75 mm x 25.25 mm. In the design process, avoid rounded edges, arcs, or similar structures, as they can significantly increase the number of nodes during mesh subdivision, leading to a decrease in simulation speed. The details are shown in Fig. \ref{fig:STL_pos}.
    
    \item \textbf{Mesh Subdivision and Boundary Condition Setup:} Use the provided script to subdivide the mesh and obtain boundary conditions as well as the marker distribution surface. Ensure that the mesh is subdivided uniformly by choosing appropriate parameters. Adjusting the initial pose is crucial, as it corresponds to subsequent rotation matrix definitions. If you do not adjust the initial pose here, many rotation matrices will need to be rewritten later, so set the initial pose at this stage.
    
    \item \textbf{Import into the Simulation Environment:} Import the designed gel part into the simulation environment. If the thickness of the gel part is large, increase the initial position to prevent overlapping between the peg and the gel during environment initialization, which can lead to mesh penetration.
    
    \item \textbf{Coordinate System Origin and Camera Parameters Adjustment:} By default, the origin of the gel's coordinate system is at the center of the gel box, and the distance from the optical center of the Gelsight Mini camera to the gel coordinate system center is 20 mm. Modify this parameter according to your team’s design. The thickness of the Gelsight Mini gel is 4 mm, so the distance from the optical center to the back of the gel should be 18 mm.
\end{enumerate}

\subsubsection{Marker Flow Generation}

When generating the marker flow, follow these steps:

\begin{enumerate}
    \item \textbf{Adjust Marker Spacing and Number:} Adjust the marker spacing and the number of markers according to your design, ensuring that the markers do not exceed the boundaries of the marker distribution surface. Ensure that the markers are evenly distributed and cover the entire sensing area.
    
    \item \textbf{Generate Marker Projection Image:} Project the markers onto a 320 x 240 resolution image. Adjust the camera's intrinsic parameters and pose to ensure that the projection meets design expectations and accurately reflects the markers’ positions in 3D space.
\end{enumerate}

These steps generate the marker flow in pixel coordinates, representing the movement of markers on the sensor surface as seen by the camera.

\section{Other Information}

\subsection{Schedule}
Here is the schedule for our challenge:

\begin{itemize}
    \item 2024-11-15: Release
    \item 2025-01-01: Stage 1 Submission Starts
    \item 2025-01-31: Stage 1 Deadline
    \item 2025-03-20: Stage 2 Deadline
    \item 2025-TBD: Winner Announcement
    \item 2025-TBD: Award Ceremony at ICRA 2025 ViTac Workshop
\end{itemize}

\subsection{Publication}
The challenge is organized as part of the \textbf{ICRA 2025 ViTac Workshop}. Participants are encouraged to submit and present their work at the workshop. Additionally, we are planning a special issue in \textbf{IEEE RAL (Robotics and Automation Letters)}. Outstanding research works from this challenge will have the opportunity to be published in this special issue.

\subsection{Award}

Table~\ref{tab:awards} shows the awards for this challenge. Each track will be evaluated independently. \textbf{The first-prize winners of each track will be invited to give a 6-minute oral presentation at the ICRA 2025 ViTac Workshop.}
\begin{table}[h]
\centering
\caption{Award Information for Each Track (All prizes in USD)}
\begin{tabular}{|l|l|l|l|}
\hline
\textbf{Prize Level} & \textbf{Track 1} & \textbf{Track 2} & \textbf{Track 3} \\
\hline
First Prize & \$2,000 × 1* & \$2,000 × 1* & \$2,000 × 1* \\
\hline
Second Prize & \$1,000 × 2 & \$1,000 × 2 & \$1,000 × 1 \\
\hline
Third Prize & \$500 × 3 & \$500 × 3 & \$500 × 1 \\
\hline
Total Pool & \$5,500 & \$5,500 & \$3,500 \\
\hline
\multicolumn{4}{l}{* First prize winners will give a 6-minute presentation at the workshop.} \\
\end{tabular}
\label{tab:awards}
\end{table}

\section{Summary and Future Work}
This paper introduces the ManiSkill-ViTac Challenge 2025, which features three key innovations:
\begin{itemize}
\item Integration of tactile and visual sensing for complex manipulation tasks
\item A novel track for sensor design optimization
\item A standardized evaluation framework bridging simulation and real-world performance
\end{itemize}

The challenge is expected to impact robotics research and industry by:
\begin{itemize}
\item Advancing tactile-vision fusion algorithms for manipulation
\item Accelerating the development of improved tactile sensor designs
\item Establishing benchmarks for contact-rich manipulation tasks
\end{itemize}

Future directions include:
\begin{itemize}
\item Expanding task complexity to multi-stage assembly operations and dexterous manipulation
\item Incorporating dynamic object interactions and deformable materials
\item Developing transfer learning strategies between simulation and real hardware
\end{itemize}

Through continued development and community engagement, we aim to foster innovations in robotic manipulation capabilities that combine multiple sensing modalities.
\bibliographystyle{IEEEtran}
\bibliography{references}

\end{document}